%% file: 0_main.tex
\documentclass[10pt,twocolumn,letterpaper]{article}

\usepackage{iccv}
\usepackage{times}
\usepackage{epsfig}
\usepackage{graphicx}
\usepackage{amsmath}
\usepackage{amssymb}

\usepackage{url}            %
\usepackage{booktabs}       %
\usepackage{amsfonts}       %
\usepackage{nicefrac}       %
\usepackage{microtype}      %
\usepackage{csquotes}
\usepackage{array}
\usepackage{adjustbox}
\usepackage{xspace}
\usepackage{multirow}
\usepackage{wrapfig}
\usepackage{import}

\usepackage{caption}
\usepackage{subcaption}
\usepackage{makecell}
\usepackage{enumitem}
\usepackage{gensymb}
\usepackage{amsmath,amssymb}
\usepackage{wrapfig}
\usepackage{ragged2e}
\usepackage{algorithm,algpseudocode}
\usepackage{color, colortbl}

\definecolor{Gray}{gray}{0.85}
\DeclareSymbolFont{largesymbolsCM}{OMX}{cmex}{m}{n}
\DeclareMathSymbol{\sumop}{\mathop}{largesymbolsCM}{"50}

\usepackage[pagebackref=true,breaklinks=true,colorlinks,bookmarks=false]{hyperref}

\def\ours{\texttt{\textbf{CoQuant}}\xspace}

\DeclareMathOperator*{\argmin}{arg\,min}

\usepackage{hyperref}

\iccvfinalcopy

\ificcvfinal\fi

\begin{document}

\setlength{\abovedisplayskip}{4pt}
\setlength{\belowdisplayskip}{4pt}

\title{Improved Techniques for Quantizing Deep Networks with Adaptive Bit-Widths}

\author{Ximeng Sun$^{1}$, Rameswar Panda$^{2,3}$, Chun-Fu (Richard) Chen$^{2,3}$, Naigang Wang$^{3}$, Bowen Pan$^{4}$, \\ Kailash Gopalakrishnan$^{3}$, Aude Oliva$^{2,4}$, Rogerio Feris$^{2,3}$, Kate Saenko$^{1,2}$ \\
$^{1}$Boston University, $^{2}$MIT-IBM Watson AI Lab, $^{3}$IBM Research, $^{4}$MIT 
}

\maketitle
\thispagestyle{empty} 

\begin{abstract}

Quantizing deep networks with adaptive bit-widths is a promising technique for efficient inference across many devices and resource constraints. In contrast to static methods that repeat the quantization process and train different models for different constraints, adaptive quantization enables us to flexibly adjust the bit-widths of a single deep network during inference for instant adaptation in different scenarios. While existing research shows encouraging results on common image classification benchmarks, this paper investigates how to train such adaptive networks more effectively. Specifically, we present two novel techniques for quantizing deep neural networks with adaptive bit-widths of weights and activations. First, we propose a collaborative strategy to choose a high-precision \enquote{teacher} for transferring knowledge to the low-precision \enquote{student} while jointly optimizing the model with all bit-widths. Second, to effectively transfer knowledge, we develop a dynamic block swapping method by randomly replacing the blocks in the lower-precision student network with the corresponding blocks in the higher-precision teacher network. Extensive experiments on multiple image classification datasets including video classification benchmarks for the first time, well demonstrate the efficacy of our approach over state-of-the-art methods.

\end{abstract} %

\subimport{./}{1_introduction}

\subimport{./}{2_relatedwork}

\subimport{./}{3_method}

\subimport{./}{4_experiments}

\subimport{./}{5_conclusion}

{\small

\input{reference.bbl}
}

\clearpage
\appendix
\subimport{./}{A1_algorithm}

\subimport{./}{A2_datasets}

\subimport{./}{A3_implementation}

\subimport{./}{A4_zeroshot}

\end{document}

%% file: 1_introduction.tex
\vspace{-12pt}
\section{Introduction} \label{sec:introduction}
\begin{figure}
\begin{center}
     \includegraphics[width=0.95\linewidth]{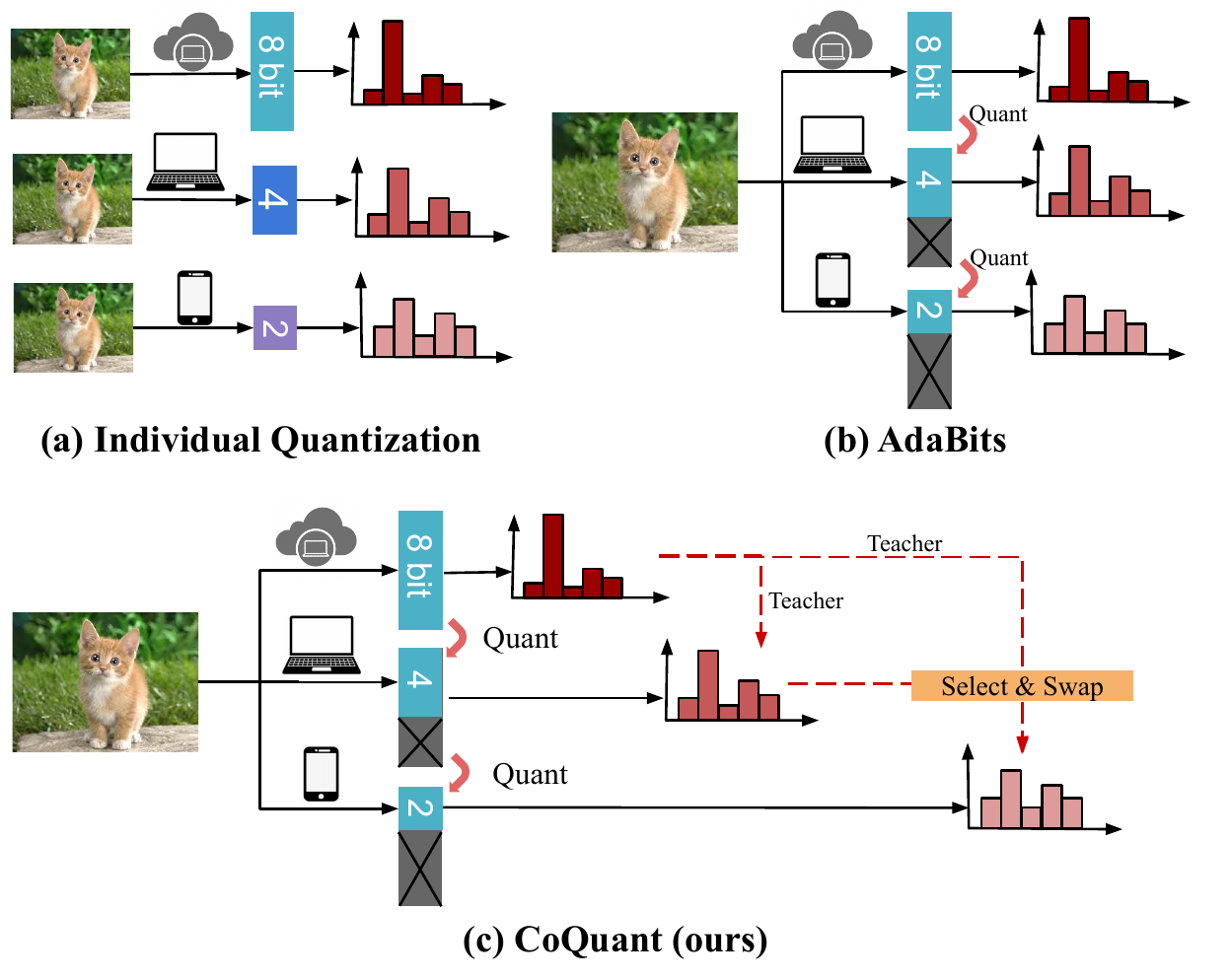}
\end{center} \vspace{-20pt}
   \caption{\small 
   \textbf{A conceptual overview of our approach}. Consider a deployment scenario with requirement of three different bit-widths (e.g., 8, 4, and 2 bits) according to  resource constraints. (a) Conventional methods train individual quantized models with specific bit-widths for each case, which prohibits instant adaptation to favor different scenarios. (b) AdaBits~\cite{jin2020adabits} jointly learns a network by treating each bit-width equally, which fails to achieve optimal performance due to increasing optimization difficulty for lower precisions. (c) We propose a novel collaborative quantization (\ours) algorithm by \textit{dynamically selecting and swapping blocks across different precisions} to transfer knowledge for effectively training a single model with all the bit-widths. %
   }
   \label{fig:concept} \vspace{-15pt}
\end{figure}

Low-precision deep networks~\cite{hubara2016binarized,zhou2016dorefa,zhang2018lq,choi2018pact,zhuang2018towards,yu2020search}, which severely reduce computation and storage by quantizing weights and activations to low-bit representations, have attracted intense attention in recent years. Despite recent progress in network quantization for improving efficiency of deep networks, most of the existing methods repeat the quantization process and retrain the low-precision network from scratch, leading to different models for different resource constraints~\cite{zhou2016dorefa,choi2018pact,zhuang2018towards} (see Figure~\ref{fig:concept}(a)). While this strategy leads to optimal efficiency for a given resource constraint, designing specialized low precision models for every practical scenario is often not flexible and economical in terms of both memory cost, and energy consumption, either with human-based methods or automatic search-based methods. 

Motivated by this, recent work~\cite{jin2020adabits} focuses on adaptive quantization which can flexibly choose the bit-width of a deep neural network during inference, to meet the dynamically changing demand. In particular, after training, we can freely quantize the weights and activations into various precision levels, without additional fine-tuning or calibration. A simple example of such network quantization is illustrated in Figure~\ref{fig:concept}(b) where the network is jointly trained under different bit-widths with shared weights. While such an approach (\eg AdaBits~\cite{jin2020adabits}) looks trivial and handy at first glance, it fails to balance the optimization difficulty across different precisions (a major challenge in quantizing deep networks with adaptive bit-widths). As a result, the higher precision tends to dominate the training, leading to sub-optimal performance across different quantization levels.

To address the above challenge, we present novel techniques for effective quantization of deep networks with adaptive bit-widths. It is well known that supervising the training of a small neural network by capturing and transferring the knowledge of a larger-parameter network can lead to a significant performance boost which is sometimes even better than that of the larger network~\cite{hinton2015distilling,cho2019efficacy,fu2020interactive,park2019relational,romero2014fitnets}.
However, unlike conventional knowledge transfer from a single teacher to a student network~\cite{hinton2015distilling,cho2019efficacy,park2019relational}, a central problem while training adaptive networks for quantization is that \textit{from which high-precision teacher to transfer knowledge given multiple teachers of different capacities}. This is an especially important problem because a very low capacity student is often unable to mimic a very strong teacher, implying the necessity of intermediate teachers to bridge the gap among them~\cite{mirzadeh2020improved}. Moreover, once the best high-precision teacher is identified, how to effectively transfer knowledge to the low-precision student is also crucial for training a single quantized neural network adaptive to diﬀerent resource constraints.   

To this end, we propose a collaborative quantization algorithm (\ours) for transferring knowledge from higher precision to lower precision while jointly optimizing the model with all the bit-widths (see Figure~\ref{fig:concept}(c)). 
Specifically, we first develop a simple yet effective teacher selection strategy to choose the best teacher adaptively to the current input by balancing the confidence of prediction and the distance in the model space. We then adopt a dynamic block swapping method to transfer knowledge by randomly replacing the blocks in the lower precision network with the corresponding blocks in the higher precision network. Our dynamic block swapping not only utilizes strong feature transformation ability of the high-precision teacher at different locations of the network, but also makes the gradient back-propagate more easily for the low-precision student. For each input batch throughout the training, we dynamically select the best teacher for lower precisions, and apply our block swapping mechanism to transfer knowledge from the selected teacher.

Experiments on two standard image classification datasets (CIFAR10~\cite{krizhevsky2009learning} and ImageNet~\cite{ILSVRC15}) show that our proposed techniques greatly improve the training efficacy of deep networks with adaptive bit-widths and outperform the recent state-of-the-art method, especially at the lower precisions, which is of significant practical value (e.g., about $1.7\%$ and $3.0\%$ improvement in 2-bit performance over AdaBits~\cite{jin2020adabits} on CIFAR10 and ImageNet, respectively). 
In addition to image classification, we extend our approach for quantizing video classification networks and observe that building low-precision networks for videos is relatively more challenging with a larger performance drop at the lower-precisions compared to image classification. Through experiments on two video classification benchmark datasets (ActivityNet~\cite{caba2015activitynet} and Mini-Kinetics~\cite{carreira2017quo}), we show that our approach achieves a significant improvement of $7\%$ in 2-bit performance over the recent method~\cite{jin2020adabits}.  
To best of our knowledge, this is the first work to report the performance of low-precision networks for action recognition in videos.

%% file: 2_relatedwork.tex
\section{Related Work} \label{sec:relatedwork}

\noindent\textbf{Network Quantization.} Several methods for quantizing deep neural networks have been studied, including binary~\cite{hubara2016binarized,rastegari2016xnor} or ternary networks~\cite{li2016ternary}, uniform quantization that uses identical bit-width for all layers~\cite{zhang2018lq,choi2018pact,zhou2016dorefa,cai2017deep,park2017weighted}, or mixed precision quantization that uses different bit-widths for different layers or even channels~\cite{wang2019haq,cai2020rethinking,wu2018mixed,yu2020search}. Designing efficient strategies for training low bit-width models using distillation~\cite{zhuang2018towards,kim2019qkd} or auxiliary module~\cite{zhuang2020training} is also another recent trend in quantization. Most relevant to our approach is AdaBits~\cite{jin2020adabits} that uses joint learning with separate clipping level parameters for training a single network with adaptive bit-widths. Despite separate clipping parameters, Adabits suffers from the performance degradation at low precision due to its interference with high precision networks which potentially disturbs the whole optimization of the network. Our approach on the other hand focuses on improving the training efficacy of such networks and utilizes a collaborative mechanism for transferring knowledge across different precision networks, leading to higher performance at lower precision. Another distinctive feature of our approach is in swapping low precision blocks with the corresponding high precision blocks which helps in easily propagating gradient through the low-precision network. 

\begin{figure*} [t]
\begin{center}
     \includegraphics[width=0.98\linewidth]{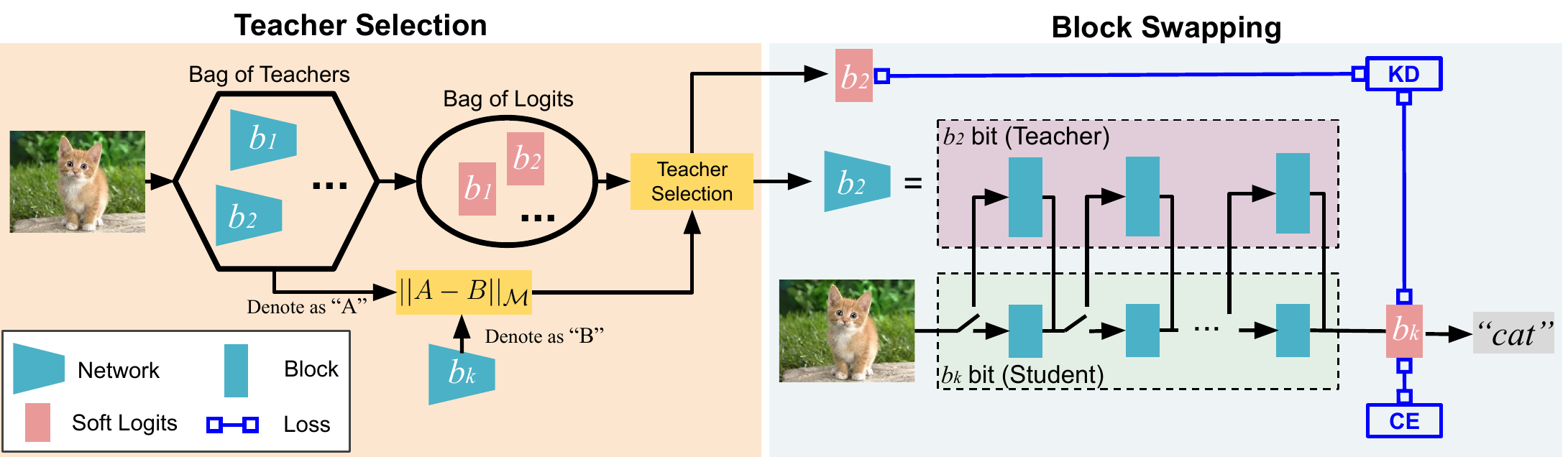}
\end{center} \vspace{-12pt}
   \caption{\small 
   \textbf{Illustration of our proposed approach.} Our approach (\ours) for training deep networks jointly with all the given precisions using a single set of weights, mainly consists of two components. First, for bit-width $b_k$, we propose a dynamic teacher selection mechanism, based on the prediction confidence and the model distance, to select the best teacher for the current input from the existing higher precisions (with bit-width $b_1, b_2, \cdots b_{k-1}$). Second, to effectively transfer the teacher's knowledge, we introduce a dynamic block swapping strategy by randomly replacing the blocks in the lower precision network with the corresponding blocks in the higher precision network. We use task-specific supervision (\eg Cross-Entropy (CE)) and a distillation loss across the selected teacher and student in training. %
   }
   \label{fig:propose} \vspace{-5pt}
\end{figure*}

\noindent\textbf{Collaborative Learning.} Collaborative learning which improves the generalization ability of a network by training with a cohort of other networks has been studied from multiple perspectives. Representative methods use knowledge distillation~\cite{hinton2015distilling,cho2019efficacy,fu2020interactive,park2019relational} for improving image classification~\cite{zhang2018deep,guo2020online}, force student networks to maintain their diversity via co-distillation~\cite{anil2018large}, or jointly train multiple network branches to establish a strong teacher~\cite{zhu2018knowledge,song2018collaborative}. %
While our approach is inspired by these methods, in this paper, we focus on collaborative learning for network quantization, where our goal is to dynamically adjust the precision of a single deep neural network without requiring additional re-training.

\noindent\textbf{Adaptive Neural Networks.} Many variants of adaptive neural networks have been recently proposed with the goal of achieving instant adaptation for different applications~\cite{bengio2015conditional,wu2018blockdrop,gao2018dynamic,kim2018nestednet}. While MSDNet~\cite{huang2017multi} 
makes early predictions to meet varying resource demands, MutualNet~\cite{yang2020mutualnet} trains a single network 
to achieve accuracy-efficiency trade-offs at runtime. Slimmable networks~\cite{yu2018slimmable} train a model to support multiple width multipliers. OFA networks adjust width, depth and kernel sizes simultaneously, to achieve better accuracy under the same computational constraints~\cite{cai2019once}. Despite recent progress, adaptive networks that are flexible to any numerical precision during inference still remains as a challenging and largely under-addressed problem.

%% file: 3_method.tex
\section{Proposed Method} \label{sec:proposedmethod}

Given a set of $n$ different bit-widths $\mathcal{B} = \{b_1, b_2, \cdots, b_n\}$ (assuming $b_1 > b_2 > \cdots >b_n$), the goal of adaptive network quantization is to seek a single set of weights, which can be executed with any bit-width $b \in \mathcal{B}$ during inference based on our demand (concerning computational resources, energy, storage, etc.), and achieves the best overall performance on $\mathcal{B}$. We denote a full precision model as $\mathcal{M}$ and its $b$-bit quantization as $\mathcal{M}_b$.  

\subsection{Preliminaries} \label{sec:prelim}

In 2D Convolution, we denote the network weights and activations by two 4D tensors $\bf{W}$ and $\mathbf{A} $.  Given a certain precision with bit-width $b$ and a quantization function $Q$, we denote the quantization of $\mathbf{W}$ and $\mathbf{A}$ as $Q(\mathbf{W}, b) = \widehat{W}_b$ and  $Q(\mathbf{A}, b) = \widehat{A}_b$.
In this paper, we adopt DoReFa~\cite{zhou2016dorefa} for weight quantization and PACT~\cite{choi2018pact}, a learnable uniform quantization scheme, for activation quantization. Note that our approach is agnostic to the type of quantization scheme and hence can work on all categories of methods.

\vspace{1mm}
\noindent\textbf{Weight Quantization.} We first normalize $\mathbf{W}$ into $[-1, 1]$ and then round it to the nearest quantization levels, as in~\cite{zhou2016dorefa}:
\begin{align}
    & \widehat{W}_b = 2 \times \text{quantize}_b(\frac{tanh(\mathbf{W})}{2 \max tanh(\mathbf{W})} + \frac{1}{2}) -1, \\
    & \text{quantize}_b(x) = \frac{1}{2^b -1} \times \lfloor (2^b - 1) x\rceil, \label{eq:old_quantize}
\end{align}
where $\lfloor . \rceil$ is the rounding operation. 

\vspace{1mm}
\noindent\textbf{Modified Weight Quantization.} 
In DoReFa~\cite{zhou2016dorefa}, all numerical precisions need to be quantized down from its 32-bit full-precision value. Even though highest precision is $b_1$ bit during inference, DoReFa requires the device to store the full-precision model $\mathcal{M}$ rather than $\widehat{M}_{b_1}$, which is memory inefficient.
To alleviate this, we modify Eq.~\ref{eq:old_quantize} to directly get $\widehat{W}_b$ from $\widehat{W}_{b_1}$ by discarding last $b_1-b$ bits in $\widehat{W}_{b_1}$ instead of re-quantizing from $W$. We further align $\mathbb{E}[\widehat{W}_b]$ with $\mathbb{E}[\widehat{W}_{b_1}]$ to minimize mean discrepancy caused by the discarded bits. 

\vspace{1mm}
\noindent\textbf{Activation Quantization.} For activation $\mathbf{A}$, we introduce a learnable clipping value $\alpha$ for each layer, whose value is originally in $[0, +\infty)$~\cite{choi2018pact}. The activation $\mathbf{A}$ is first clipped to $\alpha$ and then rounded to the nearest quantization levels:
\begin{align}
    & \widehat{A}_b = \alpha \times  \text{quantize}_b (\text{clip}(A, 0, \alpha) /\alpha). \nonumber
\end{align}

\vspace{1mm}
\noindent\textbf{Backward Propagation.} The quantization function is non-differentiable and hence it is impossible to directly apply back-propagation to train the network. To circumvent this issue, we adopt the \enquote{Straight-Through Estimator}~\cite{hinton2012neural, bengio2013estimating, zhou2016dorefa} to approximate gradient in the backward propagation:
\begin{align}
    \frac{\partial \mathcal{L}}{\partial r_i} = \frac{\partial \mathcal{L}}{\partial r_o} \frac{\partial r_o}{\partial r_i} =  \frac{\partial \mathcal{L}}{\partial r_o} \nonumber, 
\end{align}
\normalsize
where $\mathcal{L}$ is the loss function, $r_i$ is an arbitrary input and $r_o$ is the corresponding output of $\text{quantize}_b(r_i, b)$.

\subsection{Approach Overview}
Figure~\ref{fig:propose} illustrates an overview of our approach. Intuitively, the higher-precision network gains richer knowledge from the input and thus gives better performance due to its larger capacity, while a lower-precision network is more compact by sacrificing some performance. To take advantage of different precisions in the given $\mathcal{B}$, we propose two novel techniques for transferring knowledge from the existing higher precision ``teacher'' to the lower precision ``student'' while jointly optimizing a single network with all the bit-widths. 
Specifically, during the training, for the bit-width $b_k \in \mathcal{B}$, we first propose a mechanism to choose the best teacher from $\{b_1, b_2, \cdots, b_{k-1}\}$ adaptively to the current input by balancing the confidence of prediction and the distance in the model space (see Section~\ref{sec:teacher}). Second, we dynamically swap the blocks in the student with the corresponding blocks in the selected teacher network to better guide the training of the student (the current bit-width $b_k$) (see Section~\ref{sec:swapping}). We train the network using task-specific loss, such as cross-entropy loss, and distillation loss~\cite{hinton2015distilling} between teacher and student (see Section~\ref{sec:opt}). During inference, the trained network is directly set to different bit-widths, by truncating the least signiﬁcant bits, to support instant adaptation for different deployment scenarios.

\subsection{Dynamic Teacher Selection} \label{sec:teacher}
Taking a higher precision network as a teacher benefits the performance of lower precision networks~\cite{kim2019qkd}. A powerful teacher is expected to give confident predictions, and provide valuable knowledge in its soft logits, while the student gains the knowledge by mimicking the teacher. However, as shown in \cite{cho2019efficacy}, a teacher with very high capacity might be too good, that the student is unable to mimic it, resulting in the devastation of the whole optimization. Thus, optimal selection of high-precision teacher for improving the performance of low-precision student is crucial while training a single deep network that can be quantized at different levels.

Given higher precision (e.g., bit-width $b_1, b_2, \cdots, b_{k-1}$) as teacher candidates for bit-width $b_k \in \mathcal{B}$, we introduce two types of criterion to select the best teacher adaptive to the current input and the training progress, namely the confidence of prediction and the distance in the model space. We adopt entropy $H(\cdot)$ of the output logits to evaluate the prediction confidence. While it is generally hard to measure the distance in the model space, it is specifically easy for adaptive networks due to shared weights. Since each precision model is quantized from the same set of weights, we simply define the distance of two quantized networks as:
\begin{align}
    || \mathcal{M}_{b_i} - \mathcal{M}_{b_j}|| = \sum_{l = 1}^{L}  D(\widehat{W}_{b_i}^l, \widehat{W}_{b_j}^l ),
\end{align}
where $L$ is the number of layers in the network and $D(\cdot, \cdot)$ is the average of entry-wise $L_1$ distance of two matrices. There exists a trade-off between prediction confidence and model distance. In particular, the network with higher precision usually has larger capacity and gives more confident prediction, while it is further from the lower precision network. Therefore, we choose the best teacher for bit-width $b_k$ for the current batch via:
\begin{align}
    \argmin_{i \in \{b_1, b_2, \cdots, b_{k-1}\}} H(y_i) + \lambda || \mathcal{M}_{i} - \mathcal{M}_{b_k}|| ~\label{eq:teacher_select},
\end{align}
where $y_i$ is the soft logits (after Softmax) of the network with bit-width $b_i$ and $\lambda$ is a hyperparameter to balance the trade-off. When $\lambda \rightarrow 0$, the selection biased towards the highest precision. When $\lambda$ is significantly large, it is biased towards choosing $b_{k-1}$. 
Notably, the preference for teacher also shifts during the training. Initially, the performance of higher-precision models improves faster and our mechanism favors the higher precision. As the training goes on, the difference among different precisions is mitigated and it favors the closer precision. Therefore, our proposed dynamic teacher selection strategy adapts better to the current input and the training progress than the manually fixed teacher. We provide visualizations on teacher selection later in Section~\ref{sec:experiments}.

\subsection{Dynamic Block Swapping} \label{sec:swapping}
Once the optimal teacher is selected, we transfer the teacher's knowledge by dynamically swapping low-precision blocks with high-precision to make better use of the information contained in teacher network. Specifically, execution in each precision merely requires the different quantization of the same set of parameters without changing the network architecture. Benefiting from this, we dynamically swap in the high-precision teacher blocks so that the low-precision student actively incorporates the teacher's intermediate knowledge. Moreover, higher-precision blocks experience less inaccurate gradient approximation~\cite{zhuang2020training}. So swapping in these blocks helps alleviate the difficulty of propagating gradient and results in the fast-start of convergence when training a lower-precision student. Note that we use blocks and layers interchangeably throughout the paper. 

For the $l$-th block or layer in the network, let $\beta_l = \text{Bernoulli}(p_l) $ indicate whether the student block is executed ($\beta_l = 1$) or the teacher network is swapped in ($\beta_l = 0$):
\begin{align}
    A^{l+1} = \beta_l f(\widehat{A}_s^l, \widehat{W}_s^l) + (1- \beta_l)f(\widehat{A}_t^l, \widehat{W}_t^l),
\end{align}
where $\widehat{A}_s^l$ and  $\widehat{A}_t^l$ denotes the input activations of the student layer $f(\cdot, \widehat{W}_s^l)$ and the teacher layer $f(\cdot, \widehat{W}_t^l)$ correspondingly and $A^{l+1}$ is the output activation. 

Since top layers are less prone to the gradient issue and better taught by the soft logits than bottom layers, we prefer to train top layers more and then move to train the bottom layers. Therefore, we linearly increase the probability to execute the students with respect to the layer depth: 
\begin{align}
    p_l = \min(1,  (1 + l/L) p_1).
\end{align}
Following curriculum learning~\cite{bengio2009curriculum}, we set the initial value for $p_1$ and gradually increase it to 1 during training, i.e. it auto-regresses towards training the network with all student blocks at the end.
During inference, all the blocks are executed with the given precision without any swapping.

\subsection{Optimization} \label{sec:opt}
During training, we gather losses of all the precisions and then update the network. Denote the total loss as $\mathcal{L}$, and the loss for the precision with bit-width $b$ as $\mathcal{L}_b$, where
$\mathcal{L} = \sum_{b \in \mathcal{B}} \mathcal{L}_b$.
In classification tasks, $\mathcal{L}_b$ contains two parts: the cross-entropy (CE) loss with respect to the true label $y$ and a distillation loss computed by taking Kullback–Leibler (KL) divergence between the output $y_b$ and the soft logits $y_{t}$ (after softmax) provided by the chosen teacher except for the highest precision which only has the first part:
\begin{align}
    \text{KL}(y_t || y_b) = \sum_{i = 1}^{m} (y_t)_i \log \frac{(y_t)_i }{(y_s)_i },  \\
    \mathcal{L}_b = \text{CE}(y_b, y) +  \text{KL}(y_t || y_b),
\end{align}
where $m$ is the number of classes and $(\cdot)_i$ is the $i$-th element of the vector.
Different from conventional knowledge distillation with a well-trained teacher in advance~\cite{zhuang2018towards,kim2019qkd}, our model optimizes all the precisions jointly and collaboratively. Specifically, we learn all precisions with the same input batch and shared weights. Since all precisions are quantized from the same full-precision weights, the optimization processes for all precisions are intervened closely. We follow~\cite{jin2020adabits} and use a separate set of Batch Normalization layers and clipping level parameters for different precisions.

\input{tables/cifa10_joint_table}

%% file: tables/cifa10_joint_table.tex
\begin{table*}

\parbox[t]{.48 \linewidth}{
  \begin{center}
         \resizebox{0.96\linewidth}{!}{
        \begin{tabular}{c|c c c c | c}
            \Xhline{3\arrayrulewidth} 
            Methods & 8-bit & 6-bit & 4-bit & 2-bit &\cellcolor{Gray} $\Delta_\mathcal{B}$ $\uparrow$ \\
            \Xhline{3\arrayrulewidth} 
            Individual Quant. & 95.1 & 95.4 & 95.0 & 94.1 &\cellcolor{Gray} 100 \\
            \hline
            Direct Quant. (32 bit) & 10.7 & 10.4 & 10.2  & 10.8 &\cellcolor{Gray} 11.1 \\
            Direct Quant. (8 bit) & 95.1 & 93.8 & 29.9 & 8.3 &\cellcolor{Gray} 59.7 \\
            Direct Quant. (32 bit) + BN Calib & 95.1 & 94.9 & 93.0 & 9.6 &\cellcolor{Gray} 76.9 \\
            Direct Quant. (8 bit) + BN Calib & 95.1 & 94.3 & 93.1 & 31.0 &\cellcolor{Gray} 82.5 \\
            \hline
            Joint Training & 34.3 & 44.7 & 56.0 & 26.7 &\cellcolor{Gray} 42.6 \\
            Switchable BN & 94.6 & 94.5 & 94.5 & 92.3 &\cellcolor{Gray} \underline{99.2} \\
            AdaBits (CVPR'20) & 94.4 & 94.2 & 94.2 & 92.4 &\cellcolor{Gray} 99.0 \\
            \ours (Ours) & 95.2 & 95.4 & 95.1 & 94.1 &\cellcolor{Gray} \textbf{100.1}\\
            \Xhline{3\arrayrulewidth} 
        \end{tabular}
        } \vspace{-8pt}
\caption{\small \textbf{ResNet18 on CIFAR10.} \ours achieves 
best overall performance $\Delta_\mathcal{B}$ (higher is better) among all compared methods.}~\label{table:cifar10_resnet18_main}
\end{center}}
 \hfill
\parbox[t]{.47\linewidth}{
\begin{center}
\resizebox{0.96\linewidth}{!}{
        \begin{tabular}{c|c c c c | c}
            \Xhline{3\arrayrulewidth} 
            Methods & 8-bit & 6-bit & 4-bit & 2-bit &\cellcolor{Gray} $\Delta_\mathcal{B}$ $\uparrow$ \\
            \Xhline{3\arrayrulewidth} 
            Individual Quant. & 94.2 & 93.8  & 93.6 & 89.0 &\cellcolor{Gray}  100\\
            \hline
            Direct Quant. (32 bit) & 10.0 &. 9.9 & 10.0 & 10.1 &\cellcolor{Gray} 10.8 \\
            Direct Quant. (8 bit) & 94.2 & 93.0 & 80.3 & 10.9 &\cellcolor{Gray} 74.3 \\
            Direct Quant. (32 bit) + BN Calib & 91.4 & 91.3 & 89.8 & 38.2 &\cellcolor{Gray} 83.3\\
            Direct Quant. (8 bit) + BN Calib & 94.2 & 94.1 & 93.7 & 77.6 &\cellcolor{Gray} 96.9 \\
            \hline
            Joint Training & 14.2 & 15.3 & 29.2 & 46.1 &\cellcolor{Gray} 28.6 \\
            Switchable BN & 94.2 & 94.0 & 93.3 & 85.4 &\cellcolor{Gray}  99.0 \\
            AdaBits (CVPR'20) & 93.9 & 93.8 & 93.2 & 86.2 &\cellcolor{Gray} \underline{99.0}\\
            \ours (Ours) & 94.1 & 94.2 & 94.0 & 87.5 &\cellcolor{Gray}  \textbf{99.8} \\
            \Xhline{3\arrayrulewidth} 
        \end{tabular}} \vspace{-8pt}
        \caption{\small \textbf{MobileNet V2 on CIFAR10.} \ours achieves 
        best overall performance $\Delta_\mathcal{B}$ among all-at-once quantization methods.}~\label{table:cifar10_mb2_main}
\end{center}
} 
\vspace{-30pt}
\end{table*}

%% file: 4_experiments.tex
\input{tables/imagenet_minikinetics_joint_table}

\section{Experiments} \label{sec:experiments}

In this section, we conduct extensive experiments to show that our approach for effective quantization of adaptive networks outperforms the state-of-the-art method while achieving comparable performance with individual quantization models on both image and video classification datasets. 

\subsection{Experimental Setup}

\noindent\textbf{Datasets.} We evaluate the performance of our approach using several standard datasets, namely CIFAR10~\cite{krizhevsky2009learning}, ImageNet~\cite{ILSVRC15} for image classification and ActivityNet-v1.3~\cite{caba2015activitynet}, Mini-Kinetics~\cite{carreira2017quo} for action recognition in videos. ActivityNet contains 10,024 videos for training and 4,926 videos for validation across 200 action categories. Mini-Kinetics-200 (assembled by~\cite{meng2020ar}) is a subset of full Kinetics dataset~\cite{carreira2017quo} containing 121k videos for training and 10k videos for testing across 200 action classes.

\vspace{1mm}

\input{tables/activitynet_joint_table}

\begin{figure*}
     \centering
     \begin{subfigure}[b]{0.33\textwidth}
         \centering
         \includegraphics[width=0.95\textwidth]{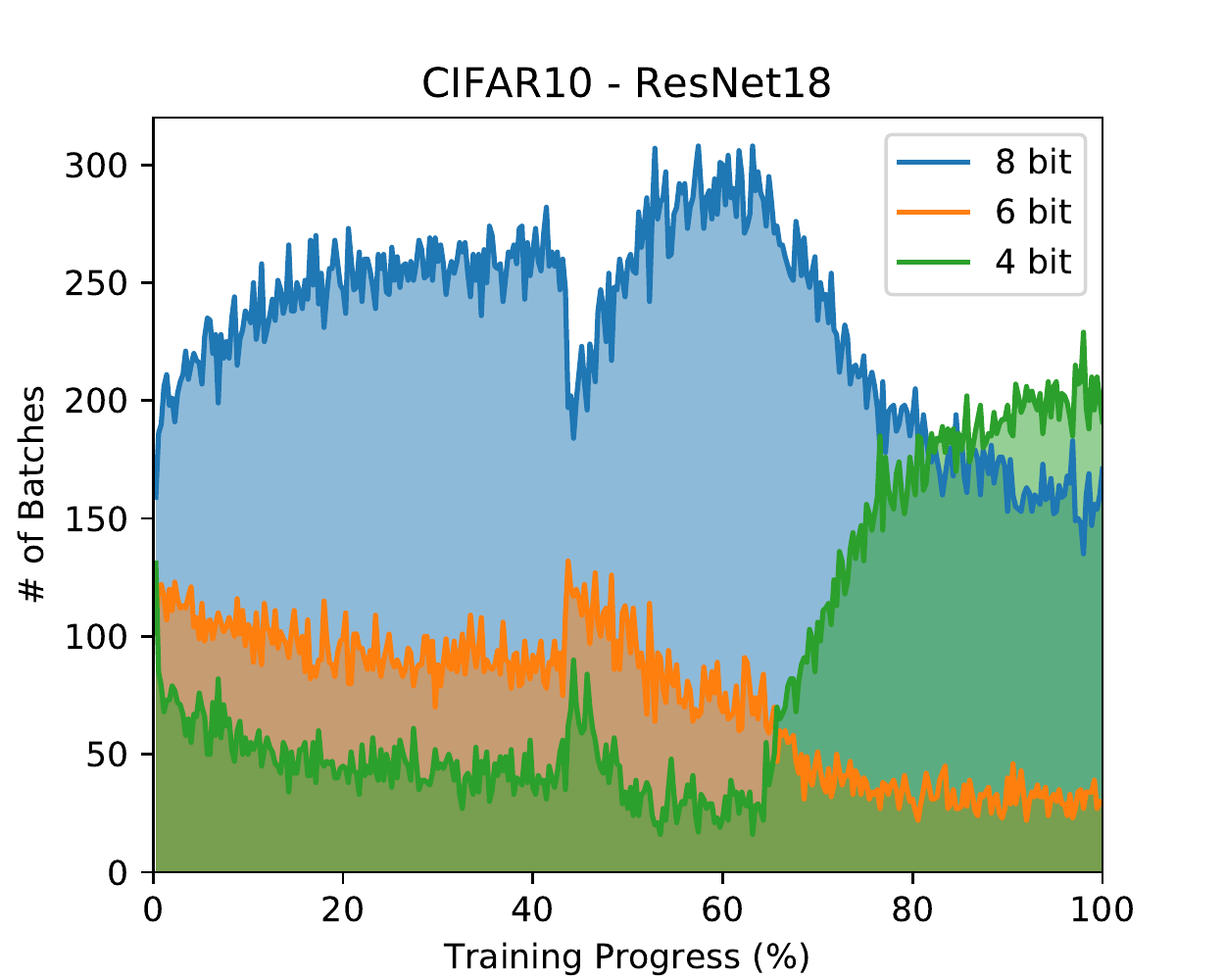}
     \end{subfigure}
     \hfill
     \begin{subfigure}[b]{0.33\textwidth}
         \centering
         \includegraphics[width=0.95\textwidth]{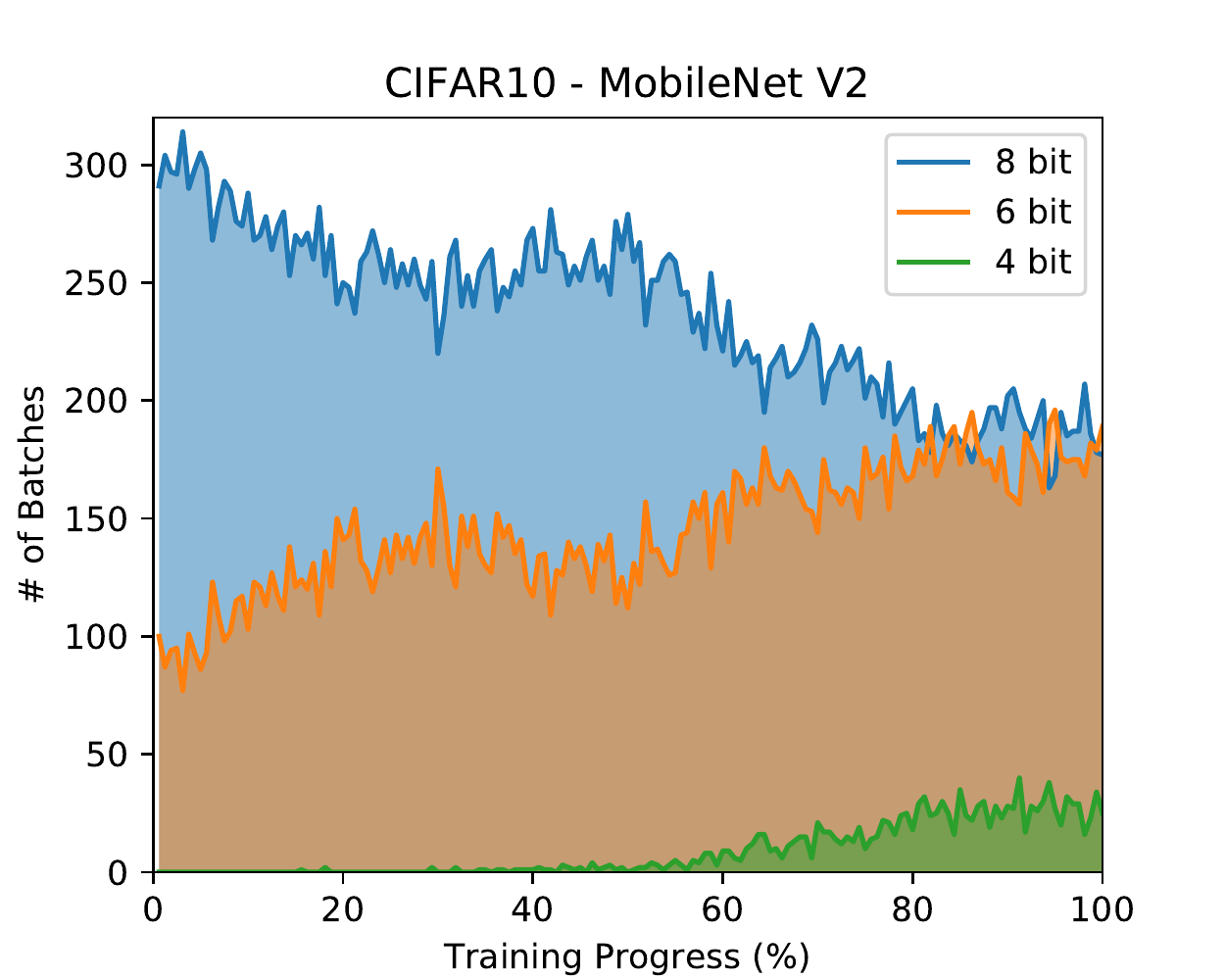}
     \end{subfigure}
     \hfill
     \begin{subfigure}[b]{0.33\textwidth}
         \centering
         \includegraphics[width=0.95\textwidth]{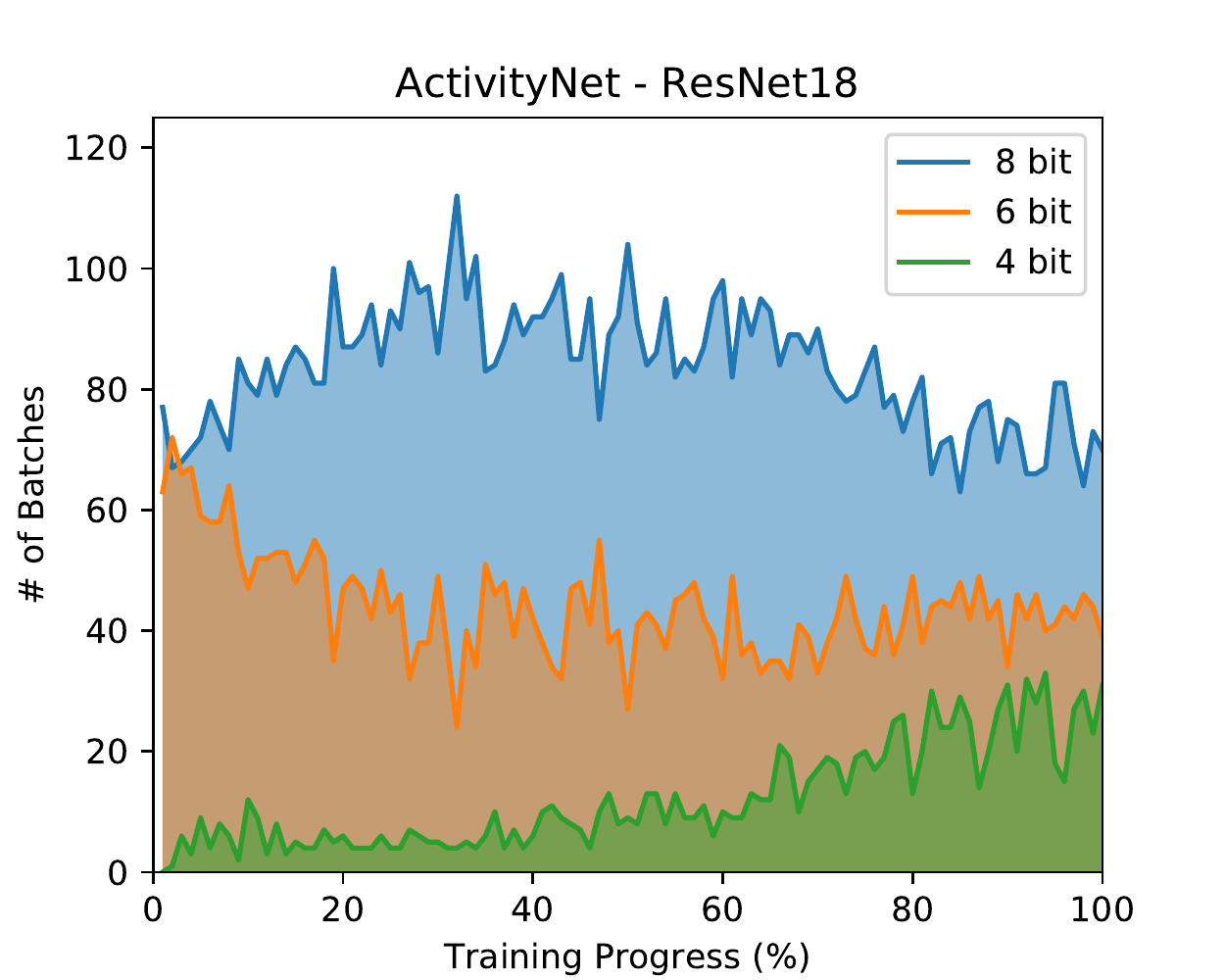}
     \end{subfigure}
     \vspace{-18pt}
        \caption{\textbf{Dynamic Teacher Selection.} We visualize the teacher selection for the 2-bit student in the training process in three scenarios. Initially, our method prefers 8-bit teacher and the preference shifts to 6-bit and even 4-bit during the training.}
        \label{fig:ts}
    \vspace{-5pt}
\end{figure*}

\noindent\textbf{Implementation Details.}
We use three different backbone architectures such as ResNet18, ResNet50~\cite{ren2015faster} and MobileNet V2~\cite{sandler2018mobilenetv2} to perform different experiments. For video classification, we adopt temporal segment network (TSN)~\cite{wang2016temporal} to aggregate the predictions over uniformly sampled 8 frames from the video. 
As shown in ~\cite{zhou2016dorefa, choi2018pact}, 8-bit model experiences no/little deficiency from full precision while 2-bit one always leads to a much worse performance. Thus, in our experiments, we set $\mathcal{B} = \{8, 6, 4, 2\}$.
For CIFAR10, we accommodate ResNet18 and MobileNet V2 
to adapt for the input size 32x32~\cite{zhang2019lookahead}.  We switch BatchNorm (BN) and clipping values for each numerical precision. We use separate sets of learning parameters (learning rate, weight decay) for clipping values of each precision. We train our network with 160, 100, 100, and 100 epochs for CIFAR10, ImageNet, ActivityNet, and Mini-Kinetics respectively. Following ~\cite{zhou2016dorefa, choi2018pact}, we do not quantize the input, the first layer and the last layer of the network. 
More implementation details are included in Appendix~\ref{sec:impl}.

\vspace{1mm}
\noindent\textbf{Baselines.} We compare our approach with the following baselines. We first consider \textbf{Individual Quantization}, where we train a separate network for each precision. We also compare with \textbf{Direct Quantization}~\cite{jin2020adabits} methods that directly quantize a higher precision network to a lower precision during inference without any extra training. We use both 32-bit full precision model and 8-bit model for testing with all precisions in $\mathcal{B}$. We also follow~\cite{yu2019network} to apply batch norm calibration (BN Calib.) to alleviate the discrepancy of the layer statistics when evaluated on another bit-width. We then compare with \textbf{Joint Training} baseline that trains a single network with all precisions simultaneously using the same BN layers and clipping values. \textbf{Switchable BN} considers the difference in layer statistics for different precisions and uses precision-specific BN layers. Finally, we compare our approach with state-of-the-art method, \textbf{AdaBits} (CVPR'20)~\cite{jin2020adabits} that applies precision-specific clipping level parameters in addition to BN layers. We use our modified DoReFa in Section~\ref{sec:prelim} and PACT~\cite{choi2018pact} as the common quantization functions in all baselines and use the same learning hyper-parameters for a fair comparison.

\vspace{1mm}
\noindent\textbf{Evaluation Metrics.} We report top-1 accuracy for all precisions on CIFAR10 and ImageNet. For video classification, we compute top-1 clip accuracy and mean Average Precision (mAP) on Mini-Kinetics and ActivityNet respectively. 
Furthermore, we report a single relative performance $\Delta_{\mathcal{B}}$ with respect to Individual Quantization to show the overall performance of different baselines as:
\begin{align}
    \Delta_{\mathcal{B}} = \frac{1}{|\mathcal{B}|}\sum_{i=1}^{|\mathcal{B}|}\frac{M_i}{M_{PACT, i}} * 100 \%, \nonumber
\end{align}
where $M_i$ and $M_{PACT, i}$ are the corresponding performance of models and individual quantization with bit-width $b_i$.

\subsection{Results and Analysis}
Table~\ref{table:cifar10_resnet18_main}-\ref{table:actn_resnet50_main} show performance in six different pairs of datasets and backbones. We separate all methods to two groups, namely direct quantization methods and methods that are specific for training adaptive quantization networks.   

\vspace{1mm}
\noindent\textbf{Image Classification.} 
We have the following key observations from Table~\ref{table:cifar10_resnet18_main}-\ref{table:imagenet_resnet18_main}.
(1) On CIFAR-10, while Individual Quantization with 2 bits reaches similar performance to 8 bits using ResNet18, there is a large gap among 8-bit and 2-bit performance on MobileNet V2, which shows that MobileNet V2, a more compact architecture, is difficult to quantize at lower bits. Similarly, the performance difference on ImageNet indicates that 2-bit quantization is very challenging on this large scale dataset, as shown in~\cite{choi2018pact,zhang2018lq,zhuang2018towards}.
(2) Direct Quantization from either 32-bit full precision or 8-bit precision experiences serious performance degradation when evaluated with a different precision far from the training one without BN Calibration~\cite{yu2018slimmable}. It is due to the mismatch of layer statistics between different precisions.
After re-calibrating BN layer parameters, Direct Quantization achieves good performance in 8 and 6 bits, especially quantized from 32-bit full precision model. However, it is still unable to recover the performance in low-precision regime, showing the necessity of joint learning approaches.
(3) When trained under all precisions, Switchable BN and AdaBits~\cite{jin2020adabits} largely improve the performance over Joint Training, which demonstrates the importance of switching BN layer parameters for different precisions in image classification. \ours consistently outperforms both of the methods on all three image classification tasks, significantly at the lower precisions, which is of great practical value. Notably, \ours improves $1.7\%$ in 2-bit performance over AdaBits for ResNet18 on CIFAR10, $1.3\%$ for MobileNet V2 on CIFAR10 and $3.0\%$ for ResNet18 on ImageNet without sacrificing the higher-precision performance. This is due to our two novel components working in concert: dynamically selecting the best high-precision teacher and then swapping low-precision blocks with the selected teacher to balance the optimization difficulty while training the adaptive quantization network with different precisions. (4) Interestingly, on CIFAR10, \ours slightly improves the higher-precision performances over Individual Quantization (Table~\ref{table:cifar10_resnet18_main} and \ref{table:cifar10_mb2_main}). We believe this is because the network easily gains information from other precisions similar to positive transfer in Multi-Task Learning~\cite{misra2016cross, vandenhende2019branched, Liu_2019_CVPR, sun2020adashare}. Overall, \ours achieves very comparable performance with Individual Quantization models on several bit-widths, while outperforming prior methods that are designed for quantizing a single deep network with different precisions.

\vspace{0.3mm}
\noindent\textbf{Video Classification.} Table~\ref{table:kinetics_resnet18_main}-\ref{table:actn_resnet50_main} summarizes the results on video classification tasks. Video datasets introduce tougher classification tasks than image datasets due to the rich temporal information present in videos and results in more diverse optimization difficulty from high to low precisions, leading to significant drop in performance from 8 to 2 bits in Individual Quantization. Similar to image classification, Direct Quantization is unable to achieve plausible performance for 2-bit model. For 2-bit performance, \ours outperforms AdaBits by a larger margin than image classification: $7.2\%$, $6.1\%$ and $7.4\%$ for ResNet18 on Mini-Kinetics, ResNet18 on ActivityNet and ResNet50 on ActivityNet respectively. To summarize, \ours improves the overall performance by about $3\%-4\%$
over AdaBits, showing its efficacy not only for image classification, but also for the challenging action classification on videos.

\vspace{0.3mm}
\noindent\textbf{Visualizations on Dynamic Teacher Selection.}
We visualize the selection of high-precision teachers for the 2-bit low-precision student throughout the training process in three different scenarios: ResNet18 on CIFAR10, MobileNet V2 on CIFAR10 and ResNet18 on ActivityNet (Figure~\ref{fig:ts}). 
We count the number of selected teachers for the input mini-batch in an epoch. From Figure~\ref{fig:ts}, we observe the shift of preference in all three cases. Specifically, our dynamic teacher selection strategy starts to favor the highest precision (8-bit) since it converges much faster at the beginning. However, as performances of 6-bit and 4-bit gradually improve during the training, it is more likely to choose the lower bit-width as the teacher for 2-bit. It is helpful since 6-bit and 4-bit models are often easier for 2-bit one to mimic.

\vspace{0.5mm}
\noindent\textbf{Zero-Shot Testing.}
\input{tables/zero_shot}
We further demonstrate the robustness of our approach \ours by evaluating the quantized model on missing bit-widths (7, 5, and 3 bits) when a network is trained with 8, 6, 4, and 2 bits. Specifically, after training with 8, 6, 4, and 2 bits, we calibrate the model to obtain BN layer parameters of the remaining 3, 5, 7 bit-widths so that the model can be executed at different precisions from 2 to 8 bits. Table~\ref{table:zero_shot} shows that \ours achieves the best performance in all the 3 remaining precisions in both cases. 
Not surprisingly, all the three methods suffer from the performance drop to some extent when executed with 3 bit compared to their 4-bit performance. Notably, on ActivityNet-ResNet50, while the 3-bit performance of both Switchable BN and AdaBits are lower than their 2-bit performance, \ours yields $66.8\%$ when evaluated with 3 bit, which is $2.2\%$ better than its 2-bit performance.

\subsection{Ablation Studies}~\label{sec:ablation}
We present the following ablation experiments using ResNet18 on CIFAR10 dataset to show the effectiveness of different components in our proposed method (Table~\ref{table:cifar10_resnet18_ablation}).

\vspace{1mm}
\noindent\textbf{Effectiveness of Dynamic Swapping.}
We implement a variant of \ours without dynamic block swapping where the selected teacher only provides  soft-logits to compute distillation loss. Better performance of our full model over \enquote{\ours w/o Swap} shows that dynamic swapping is helpful to effectively transfer knowledge and ease the optimization difficulty while training adaptive networks (see Table~\ref{table:cifar10_resnet18_ablation}).

\vspace{1mm}
\noindent\textbf{Effectiveness of Distillation Loss.} We remove the distillation loss from \ours and transfer knowledge only with dynamic block swapping. The performance of \enquote{\ours w/o DL} in Table~\ref{table:cifar10_resnet18_ablation} shows that it is important to use the teacher's soft targets in addition to dynamic swapping for better knowledge transfer across low and high precisions.

\vspace{1mm}
\noindent\textbf{Dynamic Swapping vs Feature Distillation.} We investigate the effectiveness of dynamic swapping by comparing it with intermediate feature distillation (IFD)~\cite{romero2014fitnets,zagoruyko2016paying} that also aims to transfer intermediate knowledge from a teacher to a student network. We minimize the L1 difference of the intermediate activations and observe that feature distillation (\enquote{\ours w/ IFD}) does not give competitive results with swapping. This conforms that dynamic block swapping not only transfers the intermediate knowledge but also helps in propagating gradient through low-precision networks.
\input{tables/cifar10_resnet18_ablation}

\vspace{1mm}
\noindent\textbf{Effectiveness of Dynamic Teacher Selection.} We verify the advantage of teacher selection by comparing with three methods such as one that uses the highest capacity (8-bit) network as the teacher for all lower precisions (Highest Capacity Teacher), one that uses the nearest superior bit-width as the teacher for the current precision (Recursive Teacher) and finally with random selection that randomly selects a higher precision as the teacher for each mini-batch (Random Teacher). 
Note that Highest Capacity and Recursive teacher are two extreme cases in Eq.~\ref{eq:teacher_select} by setting $\lambda$ to 0 or a significantly large value. Random teacher makes use of all candidate teachers without any preference. Table~\ref{table:cifar10_resnet18_ablation} shows that \ours outperforms all three methods, 
which demonstrates that both the predication confidence and model distance play an important role in optimal selection of teacher for transferring knowledge across different precisions. 

\vspace{1mm}
\noindent\textbf{Comparison with Ensemble Methods.}
We also compare our method with two different ensemble methods, namely Average Ensemble and Learnable Weighted Ensemble. In Average Ensemble, we simply use the average of the soft logits of multiple teacher candidates, while in Learnable Ensemble, we use a learnable linear combination of soft logits to compute the distillation loss. As seen from Table~\ref{table:cifar10_resnet18_ablation}, \ours claims better performance over two different ensemble methods, which shows that our proposed dynamic select and swap strategy makes better use of the knowledge provided by multiple teacher candidates.

\vspace{1mm}
\noindent\textbf{Comparison with Progressive Training.} Following~\cite{rusu2016progressive}, we progressively train a quantized model with multiple bit-widths in a descending order (\textit{i.e.}, from 8 bits to 2 bits by sequentially finetuning). Table~\ref{table:cifar10_resnet18_ablation} shows that it fails to preserve the performance of higher precisions and \ours outperforms it by $3.0\%$ in overall performance. In addition, we also progressively train another model with multiple bit-widths in a ascending order (from 2 bits to 8 bits) and observe that joint training under all precisions using \ours is still more effective than the ascending strategy, especially at low precision ($\sim$7\% improvement in 2-bit performance).

\input{tables/different_B}

\vspace{1mm}
\noindent\textbf{Training using Different Bit-width Combinations.} 
We also compare networks trained using different pre-defined bit-width sets to understand effect of individual bit-width on instant adaptation to different precisions. 
From Table~\ref{table:different_B}, we have the following observations. (1) If a network is trained under fewer bit-widths, the performance for the bit-width in $\mathcal{B}$ improves since it is easier for the network to handle fewer bit-widths during the training. (2) When a network is trained with $\mathcal{B} = \{8, 4\}$, 8-bit performance is much better than the one trained with $\mathcal{B} = \{8, 2\}$. It is because training with 4 bit is a closer task to training with 8 bit than training with 2 bit, which results in more positive knowledge transfer in the joint training. (3) When a network is trained with $\mathcal{B} = \{8, 6, 4, 2\}$, the middle-precisions achieve higher top-1 accuracy compared with the network trained with $\mathcal{B} = \{8, 2\}$, since the network is unable to keep the good performance for the middle precisions when the supervisions of all middle precisions are missing in $\mathcal{B}$. (4) 2 bit performance drops dramatically when the network is trained 8 and 4 bit which shows that it is very important to keep the lowest precision in $\mathcal{B}$ to achieve good performance for it during inference.

%% file: tables/imagenet_minikinetics_joint_table.tex
\begin{table*}

\parbox[t]{.48 \linewidth}{
  \begin{center}
      \resizebox{0.97\linewidth}{!}{
        \begin{tabular}{c|c c c c | c}
            \Xhline{3\arrayrulewidth} 
            Methods & 8-bit & 6-bit & 4-bit & 2-bit &\cellcolor{Gray} $\Delta_\mathcal{B}$ $\uparrow$ \\
            \Xhline{3\arrayrulewidth} 
            Individual Quant. & 69.1 & 68.8 & 68.1 & 60.1 &\cellcolor{Gray} 100 \\
           \hline
            Direct Quant. (32 bit) & 0.1 & 0.1 & 0.1 & 0.1 &\cellcolor{Gray} 0.2\\
            Direct Quant. (8 bit) & 69.1 & 13.9 & 0.1 & 0.1 &\cellcolor{Gray} 30.1 \\
            Direct Quant. (32 bit) + BN Calib & 69.3 & 68.8 & 52.3 & 0.2 &\cellcolor{Gray} 69.3 \\
            Direct Quant. (8 bit) + BN Calib & 69.1 &  68.1 & 39.3 & 0.1 &\cellcolor{Gray} 65.1 \\
            \hline
            Joint Training & 8.4 & 	11.4 & 	33.3 &  1.5 &\cellcolor{Gray} 20.0 \\
            Switchable BN & 67.9 & 	67.7 & 66.5 & 54.0 & \cellcolor{Gray} 96.0 \\
            AdaBits (CVPR'20) & 67.9 & 67.7 &  66.5 & 54.1 &\cellcolor{Gray}	\underline{96.1} \\
            \ours (Ours) & 67.9 & 67.6 & 66.6 & 57.1 & \cellcolor{Gray} \textbf{97.3} \\
            \Xhline{3\arrayrulewidth} 
        \end{tabular}}\vspace{-6pt}
 \caption{\small \textbf{ResNet18 on ImageNet.} \ours achieves 
 the best overall performance $\Delta_\mathcal{B}$ among all compared methods. }~\label{table:imagenet_resnet18_main}
\end{center}
\vspace{-12pt}
}
 \hfill
\parbox[t]{.48\linewidth}{
\begin{center}
        \resizebox{0.97\linewidth}{!}{
        \begin{tabular}{c|c c c c | c}
            \Xhline{3\arrayrulewidth} 
            Methods & 8-bit & 6-bit & 4-bit & 2-bit & \cellcolor{Gray}  $\Delta_\mathcal{B}$ $\uparrow$ \\
            \Xhline{3\arrayrulewidth} 
            Individual Quant. & 66.2 & 66.5 & 66.6 & 62.4 &\cellcolor{Gray} 100\\
            \hline
            Direct Quant. (32 bit) & 0.4 & 0.5 & 0.5 & 0.5 &\cellcolor{Gray} 0.7 \\
            Direct Quant. (8 bit) & 66.2 & 62.1 & 1.9 & 0.5 &\cellcolor{Gray} 49.3 \\
            Direct Quant. (32 bit) + BN Calib & 65.9 & 65.8 & 54.4 & 0.5 &\cellcolor{Gray}  70.3 \\
            Direct Quant. (8 bit) + BN Calib & 66.2 & 65.4 & 54.5 & 0.6 &\cellcolor{Gray} 70.3\\
            \hline
            Joint Training & 0.5 & 0.5 & 0.5 & 0.7 &\cellcolor{Gray} 0.9 \\
            Switchable BN & 64.7 & 64.8 & 63.4 & 43.9 &\cellcolor{Gray} 90.2 \\
            AdaBits (CVPR'20) & 64.1 & 64.3 & 64.2 & 48.3 &\cellcolor{Gray} \underline{91.8}\\
            \ours (Ours) & 64.6 & 64.8 & 64.4 & 55.5  &\cellcolor{Gray} \textbf{95.2} \\
            \Xhline{3\arrayrulewidth} 
        \end{tabular}} \vspace{-6pt}
    \caption{\small \textbf{ResNet18 on Mini-Kinetics.} \ours achieves
    the best overall performance $\Delta_\mathcal{B}$ among all compared methods.
    }~\label{table:kinetics_resnet18_main}

\end{center}
} 
\vspace{-35pt}
\end{table*}

%% file: tables/activitynet_joint_table.tex
\begin{table*}

\parbox[t]{.48 \linewidth}{
  \begin{center}
             \resizebox{0.97 \linewidth}{!}{
        \begin{tabular}{c|c c c c | c}
            \Xhline{3\arrayrulewidth} 
            Methods & 8-bit & 6-bit & 4-bit & 2-bit & \cellcolor{Gray} $\Delta_\mathcal{B}$ $\uparrow$ \\
            \Xhline{3\arrayrulewidth} 
            Individual Quant. & 65.3 & 65.5 & 64.3 & 59.9 & \cellcolor{Gray}  100 \\
               \hline
            Direct Quant. (32 bit) & 0.7 & 0.7 & 0.7 & 0.7 & \cellcolor{Gray} 1.1 \\
            Direct Quant. (8 bit) & 65.3 & 61.2 & 4.0 & 0.7 & \cellcolor{Gray} 50.2\\
            Direct Quant. (32 bit) + BN Calib & 67.6 & 67.5 & 57.4 & 0.7 & \cellcolor{Gray} 74.3 \\
            Direct Quant. (8 bit) + BN Calib & 65.3 & 64.7 & 56.3 & 0.9 & \cellcolor{Gray} 71.9 \\
             \hline
            Joint Training & 0.8 & 0.9 & 0.7 & 0.7 &\cellcolor{Gray} 1.2\\
            Switchable BN & 64.6 & 64.5 & 63.3 & 45.3 &\cellcolor{Gray} 92.9  \\
            AdaBits (CVPR'20) & 64.8 & 64.7 & 64.2 & 51.3 &\cellcolor{Gray} \underline{95.9}\\
            \ours (Ours) & 64.5 & 64.7 & 64.2 & 57.4 &\cellcolor{Gray} \textbf{98.3} \\
            \Xhline{3\arrayrulewidth} 
        \end{tabular}} \vspace{-6pt}
 \caption{\small\textbf{ResNet18 on ActivityNet.} \ours achieves 
 the best overall performance $\Delta_\mathcal{B}$ among all the compared methods.}~\label{table:actn_resnet18_main}
\end{center}}
 \hfill
\parbox[t]{.48\linewidth}{
\begin{center}
\ \resizebox{0.97\linewidth}{!}{
        \begin{tabular}{c|c c c c | c}
            \Xhline{3\arrayrulewidth} 
            Methods & 8-bit & 6-bit & 4-bit & 2-bit &\cellcolor{Gray} $\Delta_\mathcal{B}$ $\uparrow$ \\
            \Xhline{3\arrayrulewidth} 
            Individual Quant. &  70.0 & 69.0 & 68.9 & 64.8 &\cellcolor{Gray}\cellcolor{Gray} 100\\
               \hline
            Direct Quant. (32 bit) & 0.7 & 0.7 & 0.7 & 0.7 &\cellcolor{Gray} 1.1 \\
            Direct Quant. (8 bit) & 70.0 & 39.1 & 0.7 & 0.7 &\cellcolor{Gray}\cellcolor{Gray} 39.7 \\
            Direct Quant. (32 bit) + BN Calib & 71.5 & 71.0 & 45.8 & 0.8 &\cellcolor{Gray} 68.2 \\
            Direct Quant. (8 bit) + BN Calib & 70.0 & 66.1 & 6.2 & 0.8 &\cellcolor{Gray}  51.5 \\   
            \hline
            Joint Training & 0.7 & 0.7 & 0.8 & 0.7 &\cellcolor{Gray} 1.1\\
            Switchable BN & 68.3 & 68.3 & 67.7 & 57.1 &\cellcolor{Gray} 95.7\\
            AdaBits (CVPR'20) & 68.3 & 68.6 & 67.9 & 57.2 &\cellcolor{Gray} \underline{96.0} \\
            \ours (Ours) & 69.3 & 69.2 & 68.4 & 64.6 & \cellcolor{Gray} \textbf{99.6} \\
            \Xhline{3\arrayrulewidth} 
        \end{tabular}} \vspace{-6pt}
        \caption{\small \textbf{ResNet50 on ActivityNet.} \ours achieves 
        the best overall performance $\Delta_\mathcal{B}$ among all the compared methods.}~\label{table:actn_resnet50_main}

\end{center}
} 
\vspace{-30pt}
\end{table*}

%% file: tables/zero_shot.tex
\begin{table}
  \begin{center}
        \resizebox{0.68\linewidth}{!}{
        \begin{tabular}{c|c c c}
            \Xhline{3\arrayrulewidth} 
            Methods & 7-bit & 5-bit & 3-bit \\
            \Xhline{3\arrayrulewidth} 
             \multicolumn{4}{c}{CIFAR10 -- ResNet18} \\
             \hline
            Switchable BN &  94.4 & 94.4 & 92.4 \\
            AdaBits (CVPR'20) & 94.3 & 94.5 & 93.4  \\
            \ours (Ours) & 95.1 & 95.3 & 94.6 \\
            \hline
              \multicolumn{4}{c}{ActivityNet -- ResNet50} \\
            \hline
            Switchable BN & 68.8 &	68.5 & 56.4 \\
            AdaBits (CVPR'20) & 68.7 & 68.5 & 57.0  \\
            \ours (Ours) & 70.0 & 70.0 & 66.8 \\
            \Xhline{3\arrayrulewidth} 
        \end{tabular}}
        \vspace{-6pt}
        \caption{\small\textbf{Zero Shot Testing with BN Calib.} With ResNet18 on CIFAR10 and ResNet50 on ActivityNet, \ours achieves the best performance  when evaluating with 7, 5, and 3 bits. }~\label{table:zero_shot}
    \end{center}
\vspace{-25pt}
\end{table}

%% file: tables/cifar10_resnet18_ablation.tex
\begin{table}
  \begin{center}
        \resizebox{0.96\linewidth}{!}{
        \begin{tabular}{c|c c c c | c}
            \Xhline{3\arrayrulewidth} 
            Methods & 8-bit & 6-bit & 4-bit & 2-bit &\cellcolor{Gray} $\Delta_\mathcal{B}$ $\uparrow$ \\
            \Xhline{3\arrayrulewidth} 
            \ours w/o Swap & 94.8 & 94.8 & 94.6 & 93.2 &\cellcolor{Gray} 99.4\\
            \ours w/o DL & 94.6 & 94.5 & 94.6 & 92.7 & \cellcolor{Gray} 99.2 \\
            \ours w/ IFD & 94.7 & 94.7 & 94.7 & 93.3 & \cellcolor{Gray} 99.5 \\
            \hline
            Highest Capacity & 94.0 & 94.1 & 93.9 & 92.0 & \cellcolor{Gray} 98.6 \\
            Recursive  & 93.2 & 93.3 & 93.2 & 91.2 & \cellcolor{Gray} 97.7 \\
            Random & 94.8 & 94.8 & 94.6 & 93.2 & \cellcolor{Gray} 99.4\\
            \hline
            Average & 94.6	& 94.5 & 94.5 & 92.9 & \cellcolor{Gray} 99.2 \\
            Learnable Weighted & 94.2 & 94.2 & 94.1 & 93.0 &  \cellcolor{Gray} 99.0 \\
            \hline
            Progressive (Descend) & 91.8 & 91.9 & 92.2 & 92.9 & \cellcolor{Gray} 97.1 \\
            Progressive (Ascend) & 94.7 & 94.5 & 94.2 & 87.0 & \cellcolor{Gray} 97.6 \\
             \hline
            \ours (Ours) & 95.2 & 95.4 & 95.1 & 94.1 &\cellcolor{Gray} \textbf{100.1}\\
            \Xhline{3\arrayrulewidth} 
        \end{tabular}} \vspace{-4pt}
        \caption{\small \textbf{Ablation Study using ResNet18 on CIFAR10.} Our full model claims the best performance over all the ablated variants. }~\label{table:cifar10_resnet18_ablation}
    \end{center}
\vspace{-30pt}
\end{table}

%% file: tables/different_B.tex
\begin{table}
  \begin{center}
        \resizebox{0.68\linewidth}{!}{
        \begin{tabular}{c|c c c}
            \Xhline{3\arrayrulewidth} 
            $\mathcal{B}$ & $\{8, 6, 4, 2\}$ & $\{8, 2\}$ & $\{8, 4\}$ \\
            \Xhline{3\arrayrulewidth} 
            8-bit & 95.2 &  95.2 &  95.6 \\
            7-bit & 95.2 & 95.2 & 95.5 \\
            6-bit & 95.4 & 95.2 & 95.5 \\
            5-bit & 95.3 & 95.2 & 95.3 \\
            4-bit & 95.1 & 94.5 & 95.5  \\
            3-bit & 94.6 & 94.7 & 93.8 \\
            2-bit & 94.1 & 94.5 & 73.2\\
            \Xhline{3\arrayrulewidth} 
        \end{tabular}} \vspace{-4pt}
        \caption{\small\textbf{Evaluation Results on 2 to 8 bits with different $\mathcal{B}$.} On CIFAR10, we train networks with different pre-defined bit-width sets $\mathcal{B}$ and evaluate the performance using 2 bit to 8 bit during inference. We use zero-shot testing for all missing bit-widths.}~\label{table:different_B}
    \end{center}
    \vspace{-25pt}
\end{table}

%% file: 5_conclusion.tex
\section{Conclusion} \label{sec:conclusion}

In this paper, we present a novel collaborative knowledge transfer approach for training a single quantized network that is flexible to any numerical precision during inference without additional re-training and storing separate models. Given a low-precision student, we first choose the best high-precision teacher adaptively to the current input and then introduce a dynamic block swapping strategy to jointly optimize the model with all the precisions. 
We show the effectiveness of our approach on four standard image and video datasets, outperforming several competing methods.

%% file: A1_algorithm.tex
\section{Algorithm}
We jointly train the network under all the quantization precisions. For each mini-batch, we apply our dynamic teacher selection strategy to choose a teacher from all the higher-precision candidates and swap in the teacher block to transfer knowledge to the lower precision. We gather losses from all precisions and then update the network. We summarize our proposed method \ours for quantizing deep networks with adaptive bit-widths in Algorithm~\ref{alg:our_method}. 

\input{algorithms/our_method}

%% file: algorithms/our_method.tex
\begin{algorithm}[H]
 \caption{\ours}~\label{alg:our_method}
\begin{algorithmic}[1]
    \Require
      \Statex A bit-width set $\mathcal{B}$ for Adaptive Deep Network Quantization.
      \Statex A full-precision model $\mathcal{M}$.
    \For{$t = 1, \cdots, \text{epochs}$}
        \State Sample the current batch $(x, y)$ from the dataset.
        \State $\mathcal{L} = 0$ 
        \For{ $b \in \mathcal{B}$}
            \If{$b=b_1$}
            \State Feed-Forward: $y_b = \mathcal{M}_b (x)$
            \State $\mathcal{L}_b = \text{CE}(y_b, y)$
            \Else
            \State Choose the best teacher $\mathcal{M}_t$
            \State Feed-Forward, $y_b = \text{Swap}(\mathcal{M}_b, \mathcal{M}_t, p) (x)  $
             \State $\mathcal{L}_b = \text{CE}(y_b, y) + \text{KL}(y_t, y_b) $ 
            \EndIf
            \State $\mathcal{L} \leftarrow \mathcal{L}  +  \mathcal{L}_b $
        \EndFor
        \State Back-propagation with the loss $\mathcal{L}$
    \EndFor
  \end{algorithmic}
\end{algorithm}

%% file: A2_datasets.tex
\section{Dataset Details}
We evaluate our method using both image classification (CIFAR10~\cite{krizhevsky2009learning} and ImageNet~\cite{ILSVRC15}) and video classification (ActivityNet~\cite{caba2015activitynet} and Mini-Kinetics~\cite{carreira2017quo}) datasets. Below, we provide more details on the video classification datasets.

\vspace{1mm}
\noindent \textbf{ActivityNet.} We use ActivityNet-v1.3, an untrimmed video datatset, with 10,024 videos for training, 4926 videos for validation and 5044 videos for testing and each video has an average duration of 117 seconds. It contains 200 different daily activities (\eg drinking coffee, washing dishes, walking the dog etc.) and at least 100 untrimmed videos per class. In our paper, we train all models on the training set and test on the validation set since the test set labels are withheld by the authors. The dataset is publicly available to download at \url{http://activity-net.org/download.html}.

\input{tables/full-precision}

\vspace{1mm}
\noindent \textbf{Mini-Kinetics.} Kinetics-400 is a large-scale dataset containing 400 action classes and 240K training videos that are collected from YouTube. Since the full Kinetics dataset is quite large and the original version is no longer available from official site (about $\sim$15\% videos are missing), we use the Mini-Kinetics dataset that contains 121K videos for training and 10K videos for testing, with each video lasting 6-10 seconds. We use official training/validation splits of Mini-Kinetics released by authors~\cite{meng2020ar} in our experiments.

%% file: tables/full-precision.tex
\begin{table}
  \begin{center}
        \resizebox{0.68\linewidth}{!}{
        \begin{tabular}{c|c}
            \Xhline{3\arrayrulewidth} 
            Dataset -- Network & 32-bit \\
            \Xhline{3\arrayrulewidth} 
            CIFAR10 -- ResNet18 & 95.3 \\
            CIFAR10 -- MobileNet V2 & 94.1 \\
            ImageNet -- ResNet18 & 69.9 \\
            ActivityNet -- ResNet18 & 67.3 \\
            ActivityNet -- ResNet50 & 69.8 \\
            Mini-Kinetics -- ResNet18 & 66.3\\
            \Xhline{3\arrayrulewidth} 
        \end{tabular}}
        \caption{\small\textbf{Full-Precision Performance.} }~\label{table:full_precision}
    \end{center}
\vspace{-10pt}

\end{table}

%% file: A3_implementation.tex
\input{tables/hyperparameters}
\section{Implementation Details}
\label{sec:impl}

In this section, we provide more implementation details regarding network architectures and initialization, full-precision performance and hyper-parameter selections.  We will make our code publicly available after the acceptance.

\vspace{1mm}
\noindent \textbf{Network Architectures and Initialization.} We use the standard architectures for ImageNet, ActivityNet and Mini-Kinetics. The input image is randomly cropped to 224×224 and randomly flipped horizontally. For CIFAR10, we pad the input to 32x32 and adapt ResNet18 and MobileNet V2 as suggested~\footnote{\url{https://github.com/kuangliu/pytorch-cifar}}. We optimize 350 epochs for ResNet18 on CIFAR10 and 160 epochs for MobileNet V2 on CIFAR10. We follow~\cite{jin2020adabits} and initialize the network with the full precision model except CIFAR10 since we found that low precision models in this setting do not converge. 

\vspace{1mm}
\noindent \textbf{Full-Precision Performance.} In Table~\ref{table:full_precision}, we provide the 32-bit full-precision performance for 6 different \{dataset, architecture\} pairs that we use in the main paper to better show the performance of adaptive quantization network. 

\vspace{1mm}

\noindent \textbf{Hyper-parameters.} In Table~\ref{table:hyperparameters}, we provide hyperparameters for 6 different \{dataset, architecture\} pairs that we use in the main paper. There are 3 separate groups of hyperparameters in \ours. (1) In dynamic teacher selection, we use $\lambda$ in Eq. (4) to balance the prediction confidence and the model distance. We also define the initial value $p_1$, which is the probability of using the student block for the first layer. During the training, $p_1$ is linearly increased to 1 at the end. (2) We use a learnable uniform quantizer PACT~\cite{choi2018pact} as our activation quantizer, whose performance depends on the initial value ($\alpha_{init}$), learning rate ($\alpha_{lr}$) and the weight decay ($\alpha_{wd}$) of the clipping value $\alpha$. (3) We also provide the batch size, learning rate and learning rate scheduler that we use to train the adaptive quantization network.

%% file: tables/hyperparameters.tex
\begin{table*}
  \begin{center}
        \resizebox{0.9\linewidth}{!}{
        \begin{tabular}{c|c c| c c c| c c c}
            \Xhline{3\arrayrulewidth} 
            Dataset -- Network & init $p_1$ & $\lambda$ &  $\alpha_{init}$ & $\alpha_{lr}$ & $\alpha_{wd}$ & batch size & lr & lr scheduler\\
            \Xhline{3\arrayrulewidth} 
            CIFAR10 -- ResNet18 & 0.9 & 0.001 & 8 & 0.1 & 1e-4 & 128 & 0.1 &  MultiSteps  (150, 225, 270)  \\
            CIFAR10 -- MobileNet V2 & 0.5 & 0.001 & 8 & 0.01 & 4e-5 & 128 & 0.01 & MultiSteps (82, 122)\\
            ImageNet -- ResNet18 & 0.7 & 0.001 & 8 & 0.1 & 6e-5 & 2880 & 0.01 & MultiSteps (30, 60, 85, 95) \\
            ActivityNet -- ResNet18 & 0.5 & 0.001 & 4 & 0.01 & 5e-4 & 72 & 0.01 & Cosine\\
            ActivityNet -- ResNet50 & 0.7 & 0.0001 & 4 & 0.1 & 5e-4 & 72 & 0.01 & Cosine\\
            Mini-Kinetics -- ResNet18 & 0.7 & 0.001 & 2 & 0.1 & 1e-4 & 576 & 0.01 & Cosine \\
            \Xhline{3\arrayrulewidth} 
        \end{tabular}}
        \caption{\small\textbf{Hyperparameters.} }~\label{table:hyperparameters}
    \end{center}
\vspace{-20pt}

\end{table*}

%% file: A4_zeroshot.tex
\section{Zero Shot Testing}
\input{tables/zero_shot_others}
In the main paper, we provide zero-shot testing in two scenarios: ResNet18 on CIFAR10 and ResNet50 on ActivityNet. In this section, we compare \ours with baselines in two other scenarios (see Table~\ref{table:zero_shot_others}): ResNet 18 on ActivityNet and ResNet18 on Mini-Kinetics. In both scenarios, \ours outperforms Switchable BN and AdaBits~\cite{jin2020adabits} in all the 3 remaining quantization precisions.

%% file: tables/zero_shot_others.tex
\begin{table}
  \begin{center}
        \resizebox{0.68\linewidth}{!}{
        \begin{tabular}{c|c c c}
            \Xhline{3\arrayrulewidth} 
            Methods & 7-bit & 5-bit & 3-bit \\
            \Xhline{3\arrayrulewidth} 
             \multicolumn{4}{c}{ActivityNet -- ResNet18} \\
             \hline
             Switchable BN &  64.6 & 64.7 &  51.3 \\
            AdaBits (CVPR'20) & 64.7 & 64.5 &  53.4  \\
            \ours (Ours) & 65.5 & 65.1 & 59.3 \\
            \hline
              \multicolumn{4}{c}{Mini-Kinetics -- ResNet18} \\
            \hline
            Switchable BN &  64.2 & 64.1 & 48.7 \\
            AdaBits (CVPR'20) & 64.1 & 63.7 & 44.9  \\
            \ours (Ours) & 64.3 & 64.2 & 54.5  \\
            \Xhline{3\arrayrulewidth} 
        \end{tabular}}
        \caption{\small\textbf{Zero Shot Testing with BN Calib.} With ResNet18 on ActivityNet and ResNet18 on Mini-Kinetics, \ours achieves the best performance  when evaluating with 7, 5, and 3 bits. }~\label{table:zero_shot_others}
    \end{center}
        \vspace{-20pt}

\end{table}